\documentclass{article}
\usepackage{spconf,amsmath,graphicx}
\usepackage{xcolor}
\usepackage{hyperref}
\usepackage{amsmath}
\usepackage{multirow}
\usepackage{subfigure}
\usepackage{amsfonts}

\newcommand{\bfsection}[1]{\vspace*{0.1cm}\noindent\textbf{#1.}}

\def\tsyncnet{ModEFormer}

\title{MODEFORMER: Modality-Preserving Embedding for Audio-Video Synchronization using Transformers}
\name{Akash Gupta$^{\star}$\thanks{This work was done while Akash Gupta was at Amazon Studios.} \qquad Rohun Tripathi$^{\dagger}$ \qquad Wondong Jang$^{\dagger}$}

\address{$^{\star}$ New York University \\
$^{\dagger}$ Amazon Studios}

\begin{document}
\ninept
\maketitle
\begin{abstract}

Lack of audio-video synchronization is a common problem during television broadcasts and video conferencing, leading to an unsatisfactory viewing experience. A widely accepted paradigm is to create an error detection mechanism that identifies the cases when audio is leading or lagging. 
We propose \tsyncnet{}, which independently extracts audio and video embeddings using modality-specific transformers. 
Different from the other transformer-based approaches, \tsyncnet{} preserves the modality of the input streams which allows us to use a larger batch size with more negative audio samples for contrastive learning. Further, we propose a trade-off between the number of negative samples and number of unique samples in a batch to significantly exceed the performance of previous methods. Experimental results show that \tsyncnet{} achieves state-of-the-art performance, 94.5\% for LRS2 and 90.9\% for LRS3. Finally, we demonstrate how \tsyncnet{} can be used for offset detection for test clips.

\end{abstract}
\begin{keywords}
Contrastive learning, audio-video synchronization, transformers, negative sampling
\end{keywords}
\section{Introduction and Related Work}
\label{sec:intro}

Out-of-sync between audio and video is a critical problem that degrades user experience. This problem occurs quite often due to stochastic uncertainty from physical recording equipment or various network issues. This is more obvious in talking face videos where the lip motion does not align with the progression of audio.
Having a lip-audio sync detector is of vital importance to measure and correct offsets between the audio and video streams. 

\begin{figure}[t]
\centering
    \subfigure[SyncNet / PM]{\includegraphics[width=0.325\columnwidth]{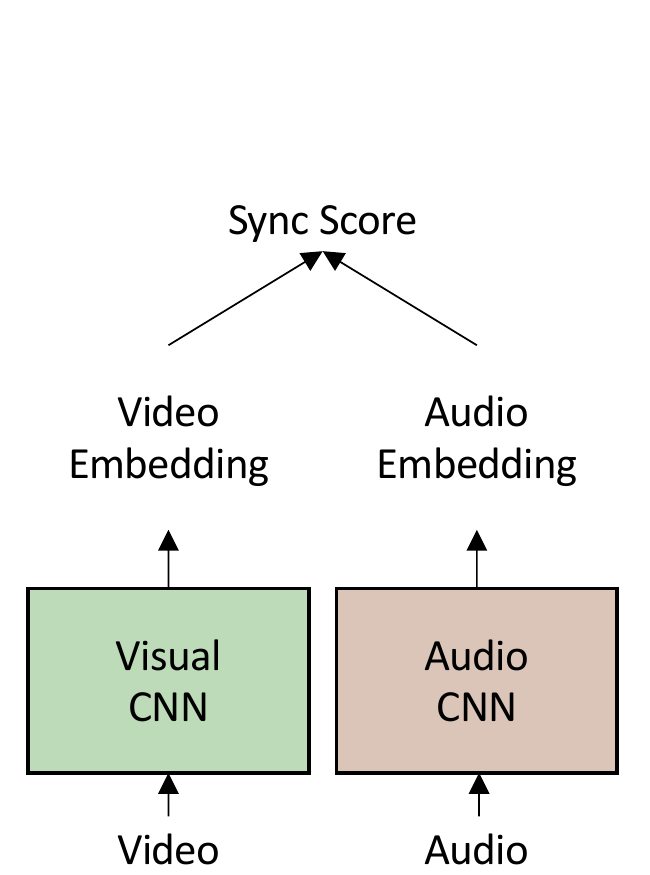}}
    \subfigure[AVST / VocaLiST]{\includegraphics[width=0.325\columnwidth]{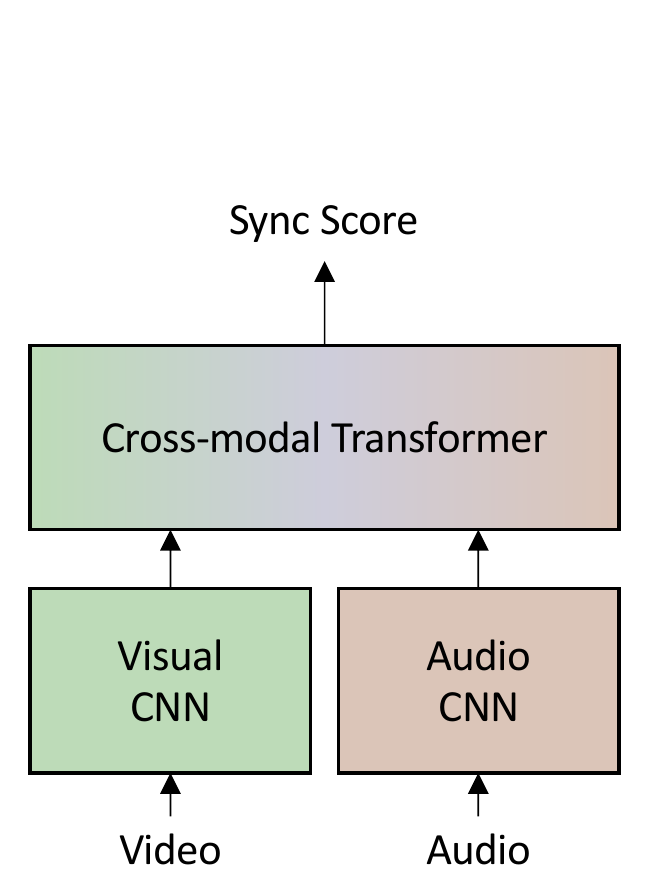}}
    \subfigure[\tsyncnet{} - Ours]{\includegraphics[width=0.325\columnwidth]{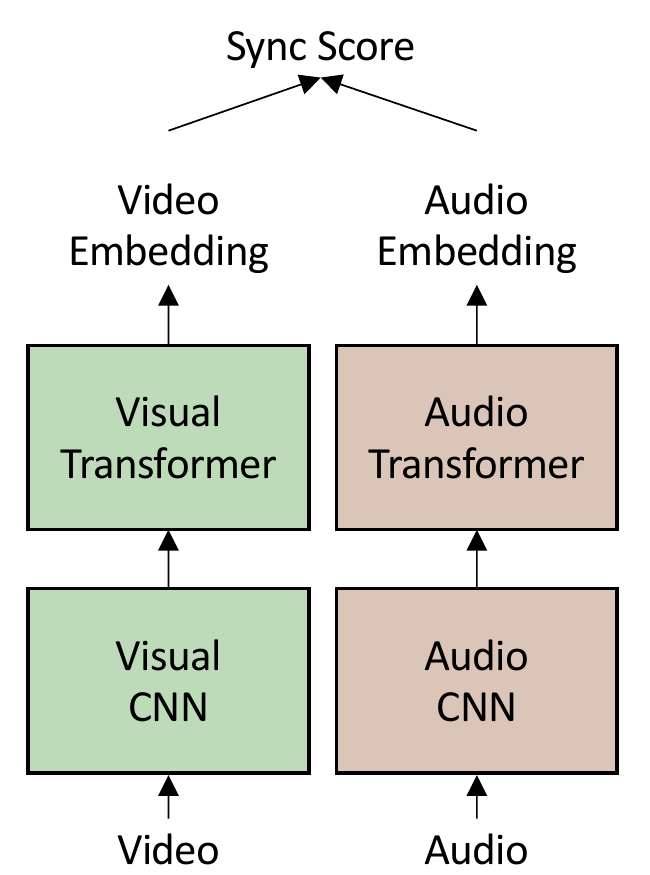}}
\caption{Comparison of different architectures for audio-video sync detection. SyncNet~\cite{Chung16a} and PM~\cite{8682524} encode audio and video inputs independently and compute a sync score by measuring cosine similarity between them. AVST~\cite{Chen21b} and VocaLiST~\cite{kadandale22_interspeech} adopt transformers to predict a sync score directly. They combine audio and visual features inside the cross-modal transformers. Our model has a modality-preserving architecture with transformers, which encodes visual and audio signals independently. By keeping the modality, we enable effective contrastive learning with either a large batch size or more negative examples.}
\label{fig:arch_comparison}
\end{figure}

Although traditional methods like time warping have proven to be quite useful in detecting this error, these are human-dependent and seem intractable with the amount of digital media in today's world. Some of the earliest works include Hershey \textit{et al.}~\cite{NIPS1999_b618c321}, which calculates the mutual information or ``synchrony'' and FaceSync \cite{NIPS2000_9f699296}, which uses the Pearson's coefficient for audio-video correlation. Other works like Morishima \textit{et al.}~\cite{5745053} and Lewis \textit{et al.}~\cite{lewis1991automated} use phoneme-viseme matching to ensure audio-video sync.

With the advancement of artificial intelligence, various sync detection techniques have been developed. SyncNet~\cite{Chung16a} is the first to use convolutional neural networks to extract audio and video embeddings and train them with a contrastive learning objective. Another following work that builds further on this is Perfect Match (PM) \cite{8682524}, which introduces the idea of multi-way matching of an audio embedding with multiple video embeddings by using a multi-way cross-entropy loss. 
They propose that this multi-way matching is beneficial during contrastive learning since it takes into account contextual information in an input sequence during training. 
Further, Prajwal \textit{et al.}~\cite{10.1145/3394171.3413532} presents a slightly different version of SyncNet by using residual convolution neural network (CNN) encoders and a cosine similarity-based loss but limits it to the binary matching of embeddings.

Transformers \cite{NIPS2017_3f5ee243} have emerged as one of the leading models for feature encoding. 
In audio-video synchronization, transformers have shown significant improvement over CNN-based architectures. Chen \textit{et al.}~\cite{Chen21b} introduced AVST or Audio-Video Sync Transformer that provides a generalized solution for synchronization over various sound classes found in in-the-wild videos. They propose a sync transformer module that helps in learning cross-modal relations between the two modalities. Another recent work, VocaLiST \cite{kadandale22_interspeech} offers an improvement over AVST by using multiple such sync transformers learning audio-to-video, video-to-audio, and hybrid correlations. Although the above modifications show improvement in audio-video synchronization, they do not explore sampling strategies for contrastive learning.

In this work, we propose \tsyncnet{}, which is a transformer-based model that extracts audio and video embeddings. We train our model to yield embeddings with high cosine similarities only when the input audio and video are in-sync. 
Unlike previous transformer-based algorithms~\cite{Chen21b, kadandale22_interspeech} that blend modalities inside their models, our method keeps separate audio and visual modalities, and thus enables contrastive learning with a large number of negative examples. Fig.~\ref{fig:arch_comparison} visually compares our \tsyncnet{}'s architecture to the existing methods~\cite{Chen21b,kadandale22_interspeech,Chung16a,8682524}. 
We conduct experiments to discover the most beneficial method of composing negative examples for contrastive learning of audio-video sync detection. We demonstrate that our method reaches state-of-the-art performance on both LRS2~\cite{8585066} and LRS3~\cite{afouras2018lrs3} datasets. The key contributions of this work are 

\begin{itemize}
  \item Building modality-preserving transformer encoders that enable contrastive learning with larger number of negative examples.
  \item Propose best trade-off between the number of negative examples and number of unique samples in a batch for optimal training of audio-video out-sync detection.

  \item Remarkable performance on LRS2 and LRS3 datasets
\end{itemize}

\section{Methodology}
\label{sec:method}

We train our model, \tsyncnet{}, to predict a sync score between a facial video and an audio clip. Our model accepts five consecutive frames ($T_\textrm{v} = 5$) of a lip region as the video input $\mathbf{v}$. We define the lip region as a lower-half of a facial video at $48 \times 96$ resolution ($H=48$ and $W=96$) by following the conventional work~\cite{kadandale22_interspeech}. We convert a given audio, whose length is corresponding to the five video frames, into a melspectrogram with 80 mel frequencies ($M=80$). We then sample $3.2$ audio frames for each video frame, and thus the input audio $\mathbf{m}$'s shape is $M\times T_\textrm{a}$, where $T_\textrm{a} = 3.2\times T_\textrm{v}= 16$. We train our model to predict a sync score between these video and audio inputs.

\subsection{Architecture of \tsyncnet{}}
\label{sec:architecture}
\begin{figure}[t]
\begin{center}
    \includegraphics[width=0.8\columnwidth]{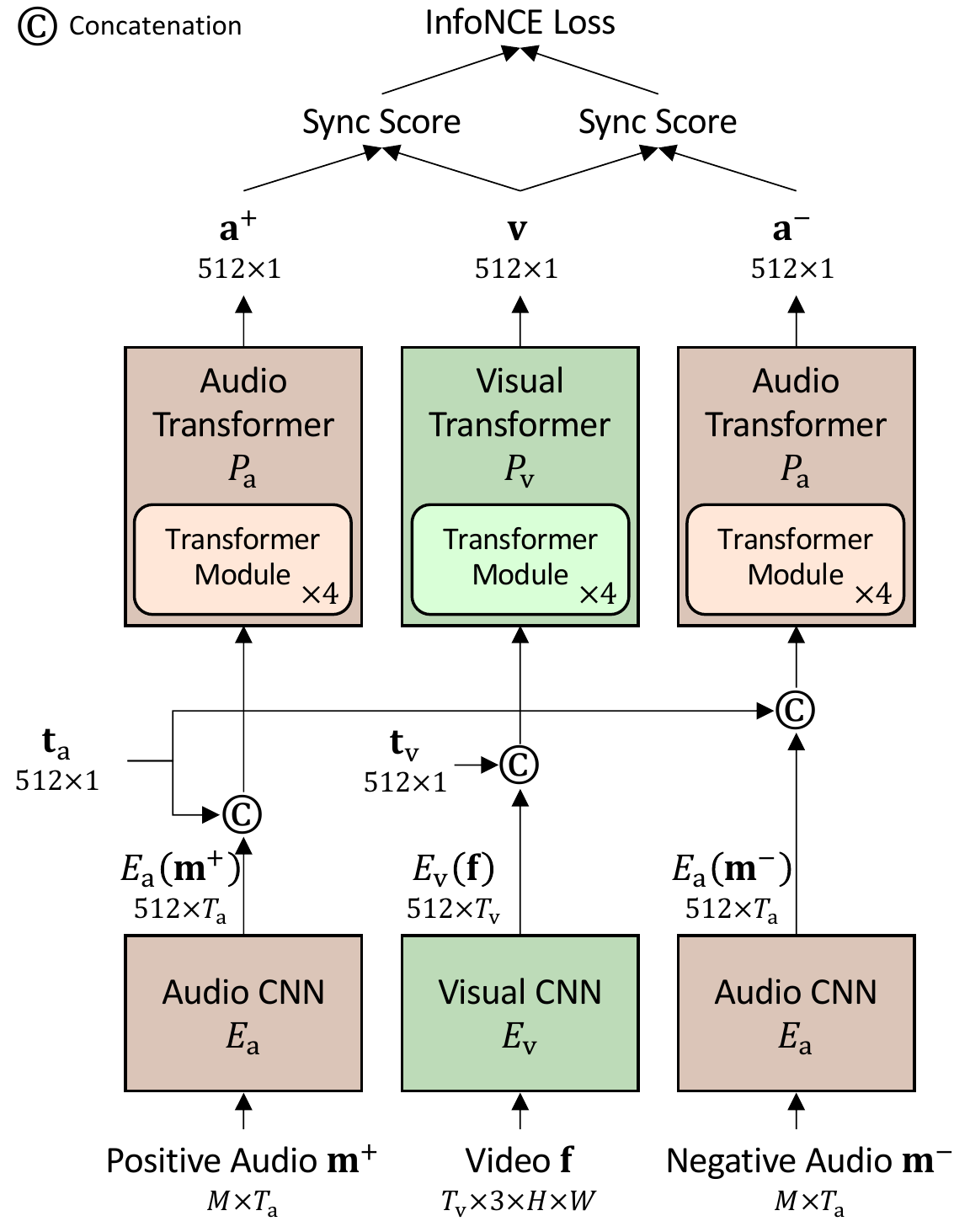}
\end{center}
\caption{Architecture of \tsyncnet{}. It takes audio and video inputs ($\mathbf{m}^-$, $\mathbf{m}^+$, $\mathbf{f}$) and yields respective embeddings ($\mathbf{a}^-$, $\mathbf{a}^+$, $\mathbf{v}$). We independently encodes the input audio and video by feeding them through the CNN and modality-specific transformer encoders ($E$ and $P$). We employ contrastive learning by minimizing an InfoNCE loss between sync scores of the embeddings.}
\label{fig:tsyncnet}
\end{figure}

We visualize the architecture of \tsyncnet{} in Fig.~\ref{fig:tsyncnet}. \tsyncnet{} consists of CNN encoders $E$ and transformer encoders $P$ for both audio and video modalities.
We design our model using modality-specific transformers and show in Sec~\ref{sec:evaluation_description} that they outperform previous transformer and CNN-based sync detectors~\cite{kadandale22_interspeech, Chen21b, Chung16a, 8682524} due to their enhanced ability of negative sampling and contrastive learning.
While the CNN encoders extract lower-level representations\footnote{Our CNN architecture differs from the one used by Petridis \textit{et al.}~\cite{8461326} as that work does not generate an individual feature map for each input frame.}, we obtain condensed modality-specific embeddings from the transformer encoders for the audio and video inputs. 

\bfsection{Lower-level representation using CNN encoders}

For the audio CNN encoder $E_\mathrm{a}$, we use an architecture similar to ResNet18~\cite{He_2016_CVPR}. We replace the first convolution layer to input a 1-channel tensor. We also adjust strides of convolutions to encode an audio input $\mathbf{m} \in \mathbb{R}^{M \times T_\mathrm{a}}$ to $E_\mathrm{a}(\mathbf{m}) \in \mathbb{R}^{512 \times T_\mathrm{a}}$ features. For the video branch, we obtain a lower-level representation $E_\mathrm{v}(\mathbf{f})\in\mathbb{R}^{512\times T_\mathrm{v}}$ by passing a video $\mathbf{f}$ through a 3D-CNN encoder $E_\mathrm{v}$ to make it learn the temporal information between frames. We adopt an architecture analogous to the audio encoder but with 3D convolutions.

\begin{table*}[t]
\centering
\caption{Comparison of audio-video sync detectors on the LRS2 and LRS3 \textit{test} sets. We measure the synchronization accuracies at different clip lengths. ``Var'' indicates whether a model has seen clips with variable lengths during training. We only compare the sync detectors trained with a fixed number of input frames. We highlight the best scores in boldface.}
\begin{tabular}{cccccccccc}
\hline
& & & \multicolumn{6}{c}{Clip Length in Frames (Seconds)} & \# of params\\
\cline{4-9}
Dataset & Model & Var & 5 (0.2s) & 7 (0.28s) & 9 (0.36s) & 11 (0.44s) & 13 (0.52s) & 15 (0.6s) & (M=Millions) \\
\hline
& \color{gray}AVST\cite{Chen21b} & \color{gray}\checkmark & \color{gray}91.9 & \color{gray}97.0 & \color{gray}98.8 & \color{gray}99.6 & \color{gray}99.8 & \color{gray}99.9 & \color{gray}42.4M\\
& SyncNet\cite{Chung16a} &  & 75.8 & 82.3 & 87.6 & 91.8 & 94.5 & 96.1 & 13.6M\\
LRS2 & PM\cite{8682524} & & 88.1 & 93.8 & 96.4 & 97.9 & 98.7 & 99.1 & 13.6M\\
& VocaLiST\cite{kadandale22_interspeech} & & 92.8 & 96.7 & 98.4 & \textbf{99.3} & 99.6 & \textbf{99.8} & 80.1M\\
& \tsyncnet{} - Ours & & \textbf{94.5} & \textbf{97.1} & \textbf{98.5} & \textbf{99.3} & \textbf{99.7} & \textbf{99.8} & 59.0M\\
\hline
\multirow{2}{*}{LRS3} & \color{gray}AVST\cite{Chen21b} & \color{gray}\checkmark & \color{gray}77.3 & \color{gray}88.0 & \color{gray}93.3 & \color{gray}96.4 & \color{gray}97.8 & \color{gray}98.6 & \color{gray}42.4M\\
 & \tsyncnet{} - Ours & & \bf 90.9 & \bf 93.1 & \bf 96.0 & \bf 97.7 & \bf 98.7 & \bf 99.2 & 59.0M\\
\hline

\hline
\end{tabular}
\label{tab:lrs_results}
\end{table*}

\bfsection{Modality-preserving encoding via transformers} 

The existing transformer-based sync detectors~\cite{Chen21b,kadandale22_interspeech} blend audio and visual features by exploiting cross-modal attentions to compute a sync score. However, combining modalities within a model makes contrastive learning difficult since a visual input should go through transformers as many times as the number of audio examples. As it requires more GPU memory, a large batch size, which is necessary for contrastive learning, is not usable. We adopt contrastive learning, which is known as effective for representing an embedding space and improving model's robustness. To this end, we preserve audio and visual modalities and compute a sync score using features from each modality unlike the previous methods. 

For an audio feature extracted from the audio CNN, we first concatenate it with an audio class token $\mathbf{t}_\mathrm{a}$, and then inject sinusoidal positional encodings~\cite{NIPS2017_3f5ee243}. 
We apply the audio transformer $P_\mathrm{a}$, which consists of four transformer modules, to the audio feature. Each transformer encoder is a lighter version of the ViT-Base architecture \cite{50650} with 4 layers. We use audio class token's output from the transformer as \tsyncnet{}'s final audio embedding $\mathbf{a}$. We process a video in the same way using the visual transformer $P_\mathrm{v}$ to obtain a final visual embedding $\mathbf{v}$.

\subsection{Negative Sampling for Contrastive Learning}
\label{sec:training}

We sample two types of negative audio examples - 1) hard negatives which are from the same video clip as the positive audio but temporally shifted. These constitute relatively similar speech features but partially or completely different phrasings. 2) Easy negatives which are from different video clips constituting different phrasings and speaker identities. We define the number of hard negatives used as $N_\mathrm{H}$. For a given batch size of $B$, each video has one positive audio and $N_\mathrm{H}$ hard negative examples. 
We train our model in two stages. In the first stage, we use $N_\mathrm{H}=2$ hard negatives for each batch entry. We do not use any easy negatives in this stage. Then in the second stage of our training for each batch entry, we use positive audios and their hard negatives in the other batch entries as easy negatives. The number of easy negatives in the second stage becomes $N_\mathrm{E}=(B-1)\times(1+N_\mathrm{H})$. 
Due to the GPU memory limit, there is a trade-off between batch size $B$ and number of hard negatives $N_\mathrm{H}$, \textit{i.e.} only small number of hard negatives are usable when a batch size is large. We study this trade-off in our experimental results to find the best ratio for contrastive learning of audio-video sync detection.

\bfsection{Loss function}
We first define a similarity $\phi$ between video embedding $\mathbf{v}$ and audio embedding $\mathbf{a}$ as the dot product between their unit vectors.
\begin{equation}
    \phi(\mathbf{v},\mathbf{a}) =\dfrac{\mathbf{v}}{|\mathbf{v}|}.\dfrac{\mathbf{a}}{|\mathbf{a}|}.
\end{equation}
We minimize the InfoNCE loss formulated as
\begin{equation}
    L = -\frac{1}{B} \sum_{\mathbf{v}, \mathbf{a}^+ \in \mathcal{P}} \log{\frac{e^{(\phi(\mathbf{v},\mathbf{a}^+)/\tau)}}{\sum_{\mathbf{a}\in \mathcal{N}(\mathbf{v})} e^{(\phi(\mathbf{v},\mathbf{a})/\tau)}}},
\end{equation}
where $\mathcal{P}$ represents a set of video and positive audio pairs and $\mathcal{N}(\mathbf{v})$ indicates a superset of positive, hard negative, and easy negative audios of video $\mathbf{v}$. 
$\tau$ is a temperature set as $0.1$. The InfoNCE loss maximizes the similarity of a video embedding with a positive audio embedding and simultaneously minimizes it with multiple negative audio embeddings.

\section{Experimental Results}

We evaluate our method on benchmark datasets, conduct study of the negative sampling for contrastive learning, and showcase our method's application as an offset detector.

\subsection{Experimental Setup}
\label{sec:evaluation_description}

We train \tsyncnet{} using the Adam optimizer with a learning rate of 0.0001. \tsyncnet{} is trained in 2 stages. In the first stage, we train the \tsyncnet{} with a batch size of $2000$ where each batch entry is from a unique clip and has two corresponding hard negative audio samples. 

In the second stage, multiple entries in the same batch are drawn from the same clip and report ablation results in Sec~\ref{sec:ablation} and Sec~\ref{sec:negative_sampling_results}

\bfsection{Datasets}
We use two public benchmark datasets for audio-video sync detection, LRS2~\cite{8585066} and LRS3~\cite{afouras2018lrs3}. Both LRS2 and LRS3 have English speakers only. LRS2 provides videos cropped around speaker's face and their corresponding audios. LRS2 consists of 96,318 \textit{pretrain}, 45,839 \textit{train}, 1,082 \textit{validation}, and 1,243 \textit{test} videos. We train \tsyncnet{} using the \textit{pretrain} split and use the \textit{validation} and \textit{test} video sets for validation and evaluation, respectively. LRS3 consists of audio-video pairs from TED and TEDx. For LRS3, both full-frames and cropped-frames are available, but we use cropped-frames only to be consistent with LRS2. LRS3 includes 1,18,516 \textit{pretrain}, 31,982 \textit{trainval}, and 1,321 \textit{test} videos. We train \tsyncnet{} on the \textit{pretrain} set and create a validation set by randomly sampling 40\% of the \textit{trainval} partition. Again, their \textit{test} partition is directly used as the test set. Please note that we follow the standard setting in the previous work~\cite{kadandale22_interspeech} to conduct fair comparisons.

\begin{figure}[t]
\centering
    \includegraphics[width=0.99\columnwidth]{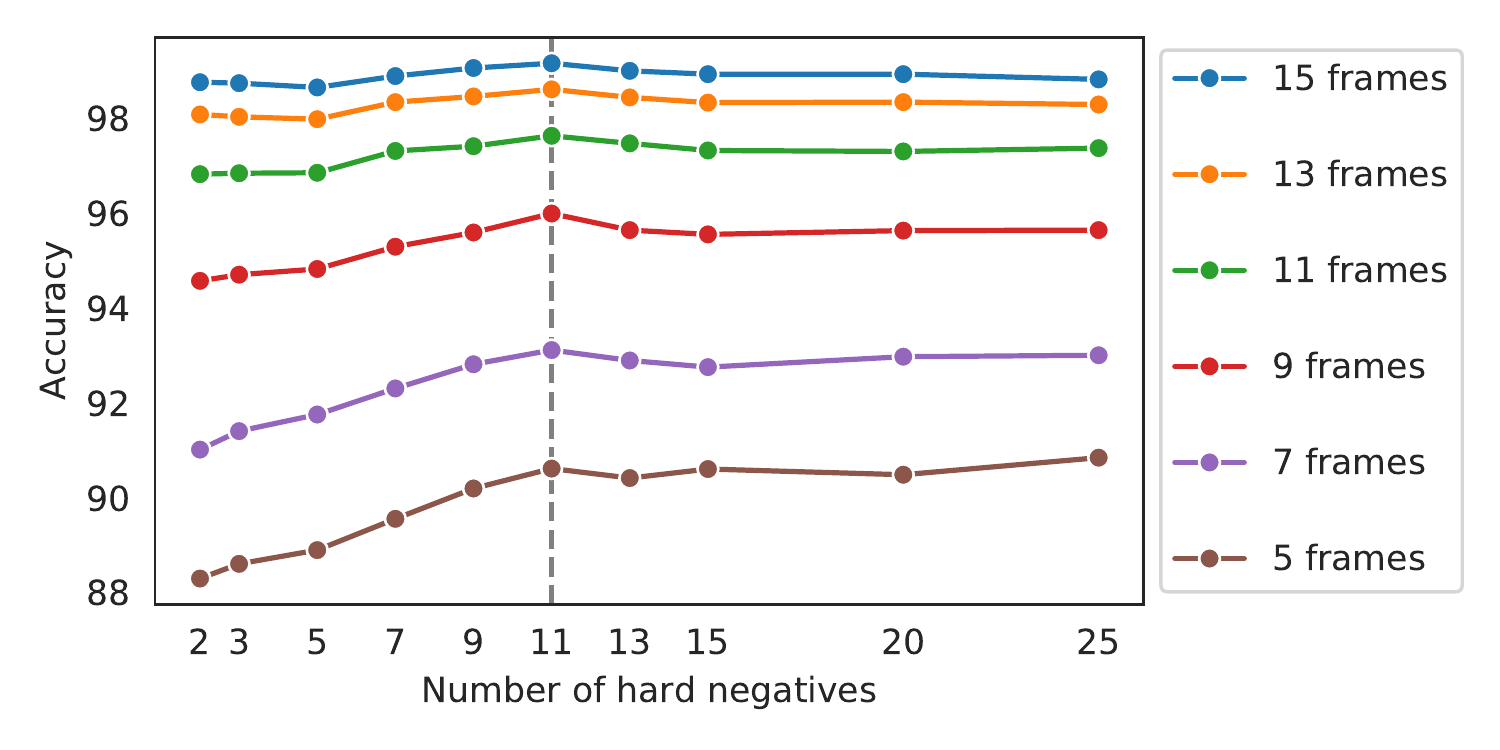}
\caption{Lip synchronization accuracies on LRS3 for different number of hard negatives. We measure the accuracy at six different clip lengths. The number of hard negatives that maximizes the overall accuracy is depicted in the grey vertical line.}
\label{fig:tradeoff}
\end{figure}

\bfsection{Evaluation metric}

We measure performances of sync detectors by the lip synchronization accuracy, which is a standard metric for benchmarking~\cite{kadandale22_interspeech}. 

For each input clip, we slide an audio window within $\pm15$ range centered on the audio with the zero offset. We determine sync detector's prediction is correct only when it gives a maximum score within $\pm1$ range.
The final lip synchronization accuracy is computed by the number of correct predictions over the number of tested clips. We report the lip synchronization accuracy for each algorithm at 6 different clip lengths which are 5, 7, 9, 11, 13, and 15.
Since our \tsyncnet{} only accepts 5 frames as the input, we slide our model over 5 frame windows when the length of the clip is longer than 5 frames and then average the cosine distance to compute the accuracy as done in the conventional works~\cite{kadandale22_interspeech}. Note that longer clip lengths lead to higher accuracy as more context is available.

\subsection{Benchmark Results}
\label{sec:benchmark_results}

Table~\ref{tab:lrs_results} lists performance of \tsyncnet{} against four different sync detectors: SyncNet~\cite{Chung16a}, PM~\cite{8682524}, AVST~\cite{Chen21b}, and VocaList~\cite{kadandale22_interspeech}.  We assess each algorithm using the lip synchronization accuracy at six different clip lengths. We observe that our \tsyncnet{} significantly and consistently outperforms all the existing methods that use a fixed number of input frames. It is even more interesting to see that our \tsyncnet{} performs better than VocaLiST~\cite{kadandale22_interspeech} considering their model (80M parameters) is heavier than ours (59M parameters). We believe the reason for such improvement is due to our modality-preserving architecture and our sampling strategy allowing us to use multiple hard negatives in contrastive learning. We do not directly compare AVST with the other sync detectors because it has seen clips of variable lengths as input during training. As can be seen in Table~\ref{tab:lrs_results}, \tsyncnet{} outperforms AVST significantly by 2.6\% on 5-frames but AVST performs better when the length of the frames is more than 7 frames with a difference of less than 0.3\%. This can be attributed to their model having the information of all frames when it makes predictions, whereas we use the sliding window strategy explained in Sec~\ref{sec:evaluation_description}.

Table~\ref{tab:lrs_results} shows synchronization accuracy of our \tsyncnet{} and AVST~\cite{Chen21b} on the LRS3 dataset. 
We present \tsyncnet{}'s performance here to demonstrate the generalizability of our approach. Our \tsyncnet{} achieves the SoTA on LRS3, which aligns with the results on LRS2. To the best of our knowledge, there are no existing works that have evaluated sync detectors on LRS3's cropped faces. AVST's accuracy is available for LRS3, but unfortunately, they use full-frame videos.

\begin{table}[t]
\centering
\caption{Results of 3D-SyncNet and \tsyncnet{} on LRS3 test set }
\begin{tabular}{cccc}
\hline
 & \multirow{2}{*}{3D-SyncNet} & \tsyncnet{} & \tsyncnet{} \\
 & & (1st stage) & (2nd stage) \\
\hline
Accuracy & 80.2\% & 88.3\% & 90.9\% \\
\hline
\end{tabular}
\label{tab:3dsyncnet_results}
\end{table}

\subsection{Architecture Ablation}
\label{sec:ablation}
Table~\ref{tab:3dsyncnet_results} compares 3D-SyncNet and \tsyncnet{} in both the training stages as described in Sec.~\ref{sec:evaluation_description}, on the LRS3 test set in terms of the lip synchronization accuracy at the clip length of five frames. We build 3D-SyncNet by ablating transformers from \tsyncnet{}. We train both the models with the same InfoNCE loss and sampling strategy. Ablation of the transformer encoders decreases the performance from 88.3\% to 80.2\%, which shows that attention helps in learning better latent representations. Further, we also see the benefit of the second stage training with more negative examples for \tsyncnet{} improving by 2.6\%.

\begin{figure}[t]
    \centering     
    \begin{tabular}{cc}
    \includegraphics[width=0.43\columnwidth]{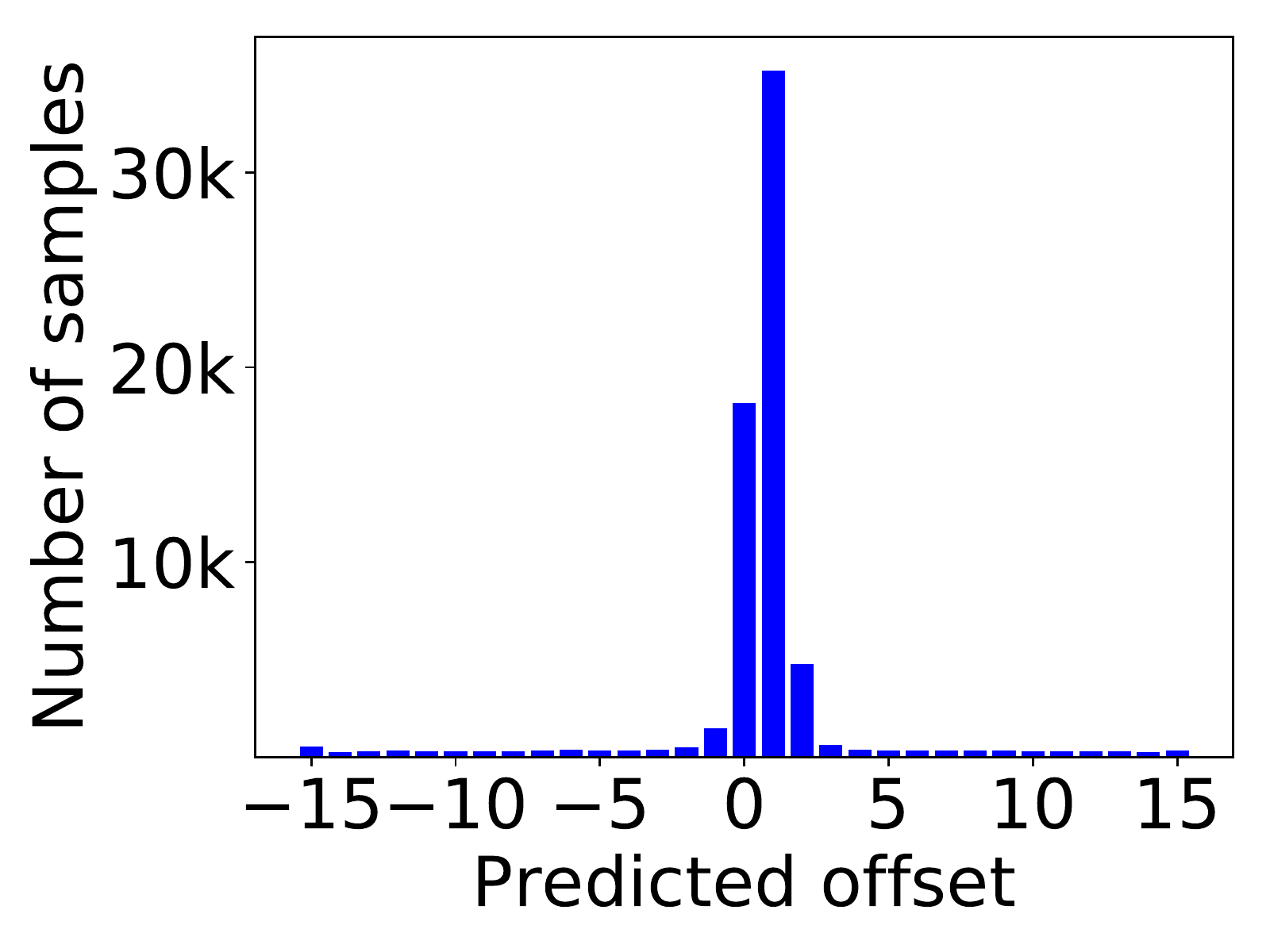} & \includegraphics[width=0.43\columnwidth]{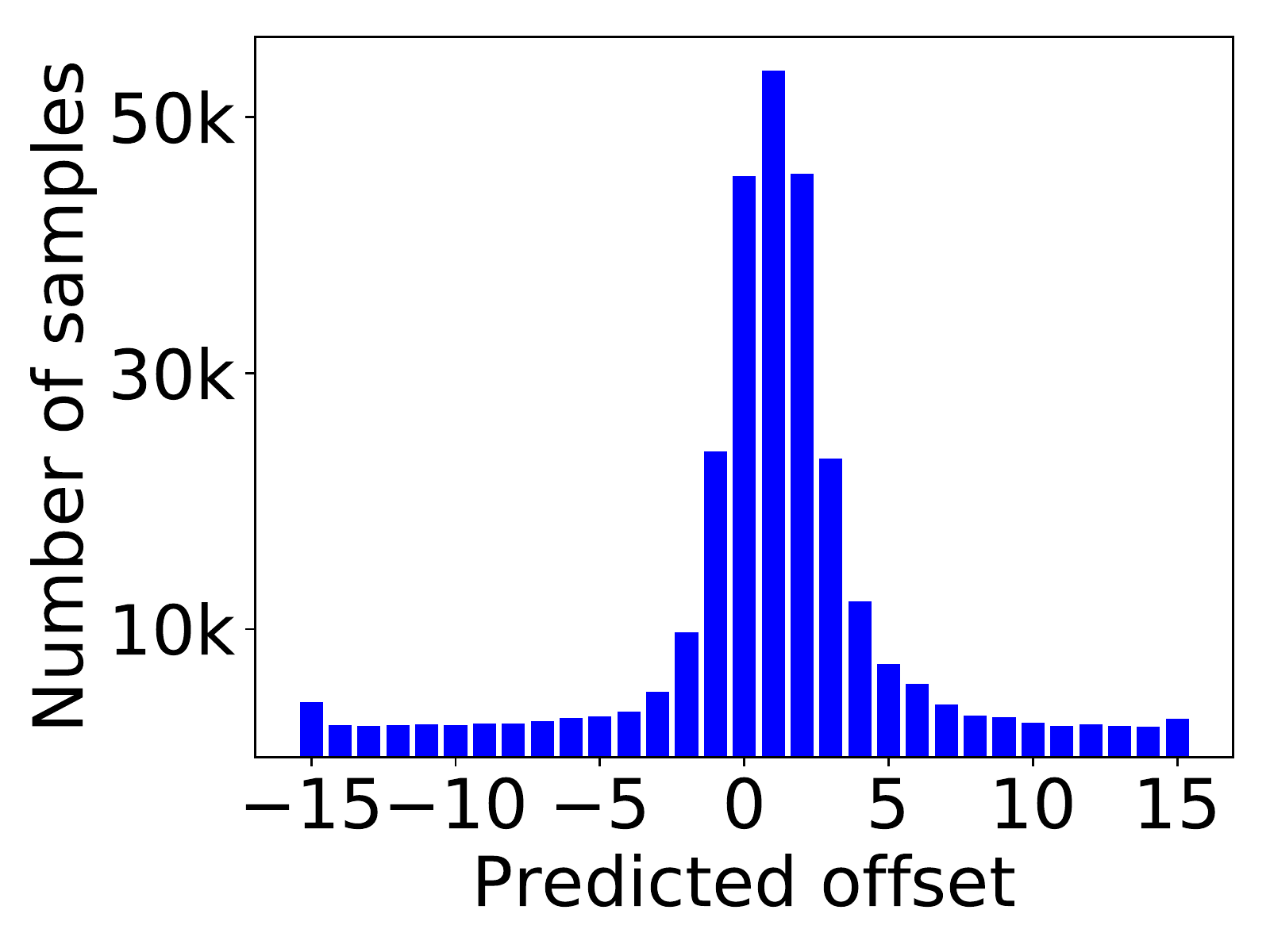}\\
    \qquad(a) & \qquad(b)\\
    \multicolumn{2}{c}{Trained on LRS2 and tested on (a) LRS3, (b) VoxCeleb2}\\
    \includegraphics[width=0.43\columnwidth]{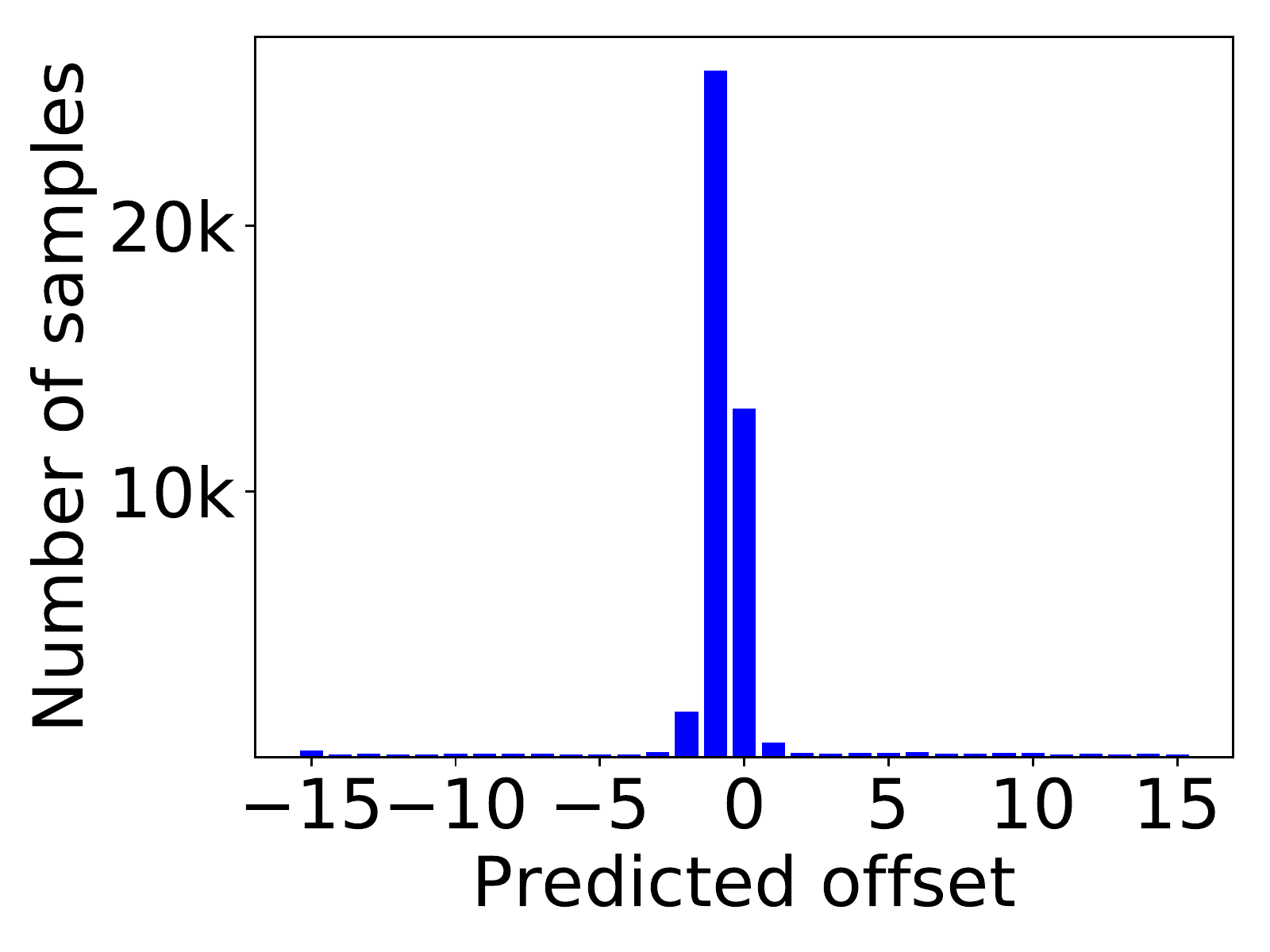} & \includegraphics[width=0.43\columnwidth]{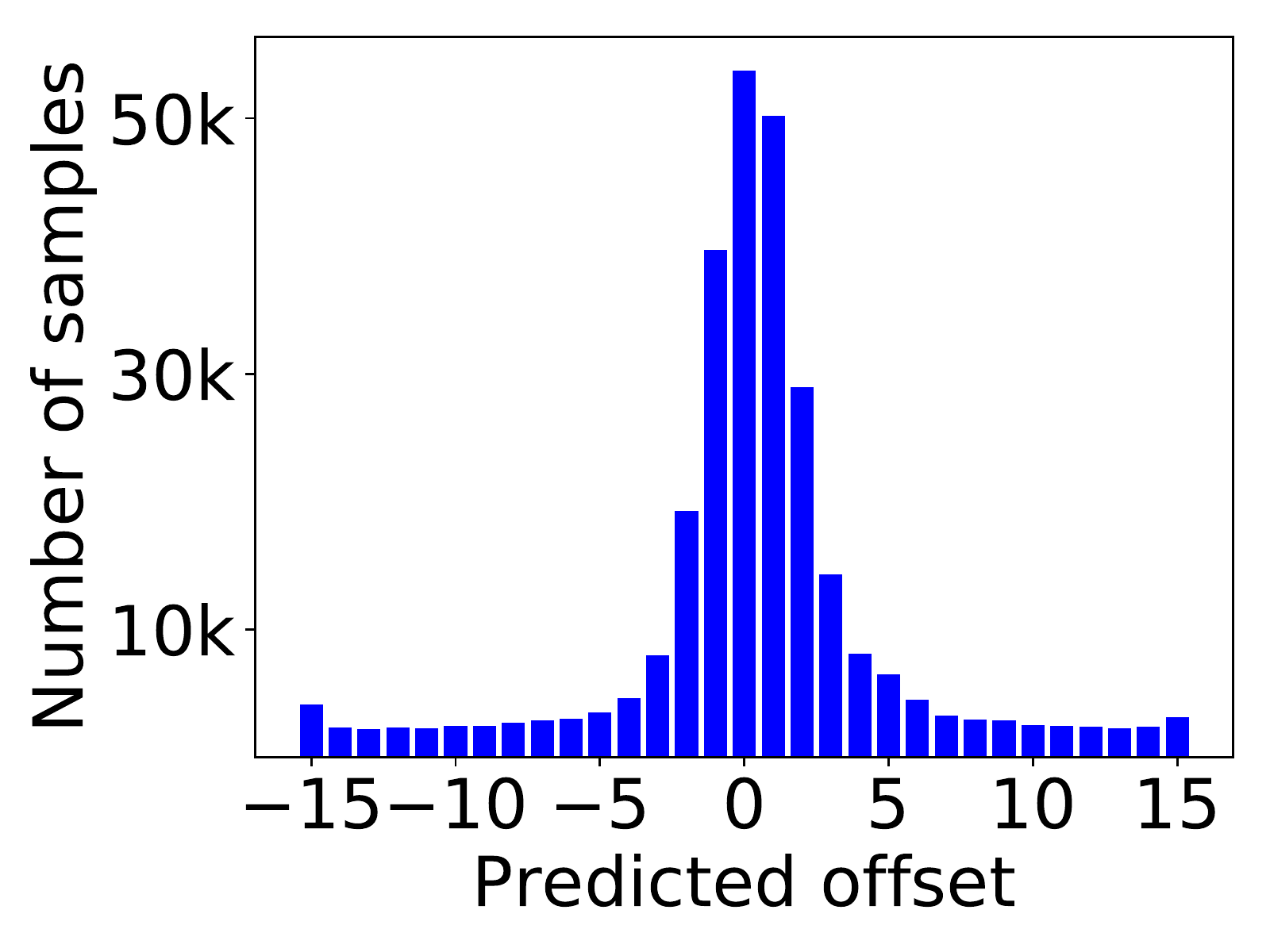}\\
    \qquad(c) & \qquad(d)\\
    \multicolumn{2}{c}{Trained on LRS3 and tested on (c) LRS2, (d) VoxCeleb2}
    \end{tabular}
    \caption{Histogram of predicted offsets for \tsyncnet{} models tested on out of distribution datasets}
    \label{fig:histogram_results}
\end{figure}

\subsection{Negative Sampling Strategy}
\label{sec:negative_sampling_results}

We experiment further to find the optimal number of hard negatives between 2 to 25 by measuring the lip synchronization accuracy. Fig.~\ref{fig:tradeoff} illustrates the accuracies for LRS3 at clip lengths of 5, 7, 9, 11, 13, and 15 frames in accordance with the number of hard negatives. We see that the overall lip sync accuracy peaks when the number of hard negatives is 11. This shows that balancing between the number of hard negatives and batch size is important to maximize performance. 

\subsection{Offset Detection}

We apply a trained \tsyncnet{} to detect any audio-video lag in a given test clip. For a given clip, we compute the cosine similarities at every video frame for audio windows in its neighborhood as described in Sec~\ref{sec:evaluation_description}. We identify the predicted offset for each video frame as the audio window with the highest cosine similarity. Then we generate the histogram of predicted offsets for each of the video frames in the test clip as shown in Fig.~\ref{fig:histogram_results}. The largest peak of the predicted offset histogram indicates the offset in the test clip. Note that this offset is in comparison to the training dataset used for training the \tsyncnet{}. 

Using this mechanism, we discovered that the LRS2 and LRS3 datasets are not in sync with one another. The largest peak of the predicted offset is $+1$ (+0.04s) when using a \tsyncnet{} trained on LRS2 and tested on the LRS3 test set, Fig.~\ref{fig:histogram_results}(a). In parallel, for a \tsyncnet{} trained on LRS3 and tested on LRS2 test set, the offset is at $-1$ (-0.04s), Fig~\ref{fig:histogram_results}(c).
To hypothesize which of the datasets is out of sync, we tested models from both LRS2 and LRS3 on an out of distribution test dataset of 1,568 random clips from VoxCeleb2~\cite{VoxCeleb2} which is a multi-lingual dataset with noisier audio than LRS2 and LRS3~\cite{asr_is_all_you_need}. 
We find that an LRS3 trained model has the largest peak of the predicted offset at $0$, Fig.~\ref{fig:histogram_results}(d) while the largest peak of the predicted offset for an LRS2 trained model is $1$, Fig.~\ref{fig:histogram_results}(b). These results lead us to hypothesize that LRS3 and VoxCeleb2 are in sync while LRS2 is out of sync.

Further, if we shift the audio windows in the LRS2 test set with an offset of $+1$, the accuracy of the LRS3 model on the test set increases from 88.27\% to 91.39\%. In future experiments, we recommend accounting for this offset when reporting performance across the datasets.

\section{Conclusion}

We present a modality-preserving sync detector, \tsyncnet{}, which yields state-of-the-art performance on both the LRS2 and LRS3 datasets. Since \tsyncnet{} preserves the modality, we are able to trade-off between the number of negative examples and the number of unique samples in a batch to find the most optimal configuration. Further, we demonstrate offset detection using \tsyncnet{} for an out-of-distribution test dataset and hypothesize that LRS2 and LRS3 are out-of-sync by +0.04 seconds

\bibliographystyle{IEEE}
\bibliography{main}

\end{document}